\documentclass[10pt,twocolumn,letterpaper]{article}

\usepackage{iccv}
\usepackage{times}
\usepackage{epsfig}
\usepackage{graphicx}
\usepackage{amsmath}
\usepackage{amssymb}

\usepackage{booktabs}
\usepackage{multirow}
\usepackage{enumitem}
\usepackage[table]{xcolor}
\usepackage{makecell}
\usepackage{xcolor}
\usepackage{algorithm}
\usepackage{algorithmic}
\usepackage{verbatim}
\usepackage{fontawesome5}
\usepackage[accsupp]{axessibility}

\usepackage[pagebackref=true,breaklinks=true,letterpaper=true,colorlinks,bookmarks=false]{hyperref}

\iccvfinalcopy % *** Uncomment this line for the final submission

\def\iccvPaperID{5840} % *** Enter the ICCV Paper ID here

\def\ie{\emph{i.e.}}
\def\eg{\emph{e.g.}}

\newcommand{\para}[1]{\vspace{.05in}\noindent\textbf{#1}}

\newcommand{\name}{ConSlide }
\ificcvfinal\fi

\begin{document}

%%%%%%%%% TITLE
\title{ConSlide: Asynchronous Hierarchical Interaction Transformer with Breakup-Reorganize Rehearsal for Continual Whole Slide Image Analysis}

\author{Yanyan Huang$^{1,2}$\footnotemark[1], Weiqin Zhao$^{1}$\footnotemark[1], Shujun Wang$^{3}$, Yu Fu$^{2}$, Yuming Jiang$^{4}$, Lequan Yu$^{1}$\footnotemark[2]\\
$^{1}$The University of Hong Kong\\
$^{2}$Zhejiang University\\
$^{3}$The Hong Kong Polytechnic University\\
$^{4}$Stanford University\\
{\tt\small \{yanyanh, wqzhao98\}@connect.hku.hk, shu-jun.wang@polyu.edu.hk,}\\
{\tt\small yufu1994@zju.edu.cn, ymjiang2@stanford.edu, lqyu@hku.hk}
% \and
}

\maketitle

\renewcommand{\thefootnote}{\fnsymbol{footnote}}
\footnotetext[1]{These authors contributed equally to this work.}
\footnotetext[2]{Corresponding Author.}
\ificcvfinal\thispagestyle{empty}\fi

%%%%%%%%% ABSTRACT
\begin{abstract}

Whole slide image (WSI) analysis has become increasingly important in the medical imaging community, enabling automated and objective diagnosis, prognosis, and therapeutic-response prediction. 
%
% However, in clinical practice, the continuous progress of evolving WSI acquisition technology, the diversity of scanners, and different imaging protocols hamper the utility of WSI analysis models.
However, in clinical practice, the ever-evolving environment hamper the utility of WSI analysis models.
%
% However, existing WSI analysis methods usually adopt the static model learning setting, with the insufficient capability of ?adapting to new tasks/datasets over time. 
%
In this paper, we propose the FIRST continual learning framework for WSI analysis, named \textbf{ConSlide}, to tackle the challenges of enormous image size, utilization of hierarchical structure, and catastrophic forgetting by progressive model updating on multiple sequential datasets. 
Our framework contains three key components. The Hierarchical Interaction Transformer (HIT) is proposed to model and utilize the hierarchical structural knowledge of WSI. The Breakup-Reorganize (BuRo) rehearsal method is developed for WSI data replay with efficient region storing buffer and WSI reorganizing operation. 
%
% Finally, during the replay stage, we devise an 
The asynchronous updating mechanism is devised to encourage the network to learn generic and specific knowledge respectively during the replay stage, based on a nested cross-scale similarity learning (CSSL) module. 
We evaluated the proposed ConSlide on four public WSI datasets from TCGA projects. It performs best over other state-of-the-art methods with a fair WSI-based continual learning setting and achieves a better trade-off of the overall performance and forgetting on previous tasks.

\end{abstract}

\section{Introduction}

Whole slide images (WSIs) contain rich histopathological information of the tissue sections and are routinely used for clinical practice.
With recent advances in deep learning, computational WSI have attracted widespread attention in the medical image analysis community, which can provide automated and objective diagnosis, prognosis, and therapeutic-response prediction~\cite{cornish2012whole,lu2021data,shao2021transmil,tanizaki2016report}. 
However, the huge size and expensive pixel-level annotations of WSI bring computational challenges in deep model architecture design~\cite{lu2020capturing,huang2022deep,javed2020multiplex}.
%Moreover, WSI datasets usually only have slide-level labels, as manual pixel-level annotations of WSI need more effort to obtain, like time, expense, and specific knowledge of pathologist~\cite{liu2017detecting,guan2022node}.
%
Therefore, multiple instance learning (MIL)-based approaches are proposed for weakly-supervised WSI analysis~\cite{campanella2019clinical,lu2021data,shmatko2022artificial}, where they first divide WSI into a set of patches, conduct analysis for each patch, and then aggregate them together for slide-level prediction.

\begin{figure}[t]
\centering
\includegraphics[width=1\columnwidth]{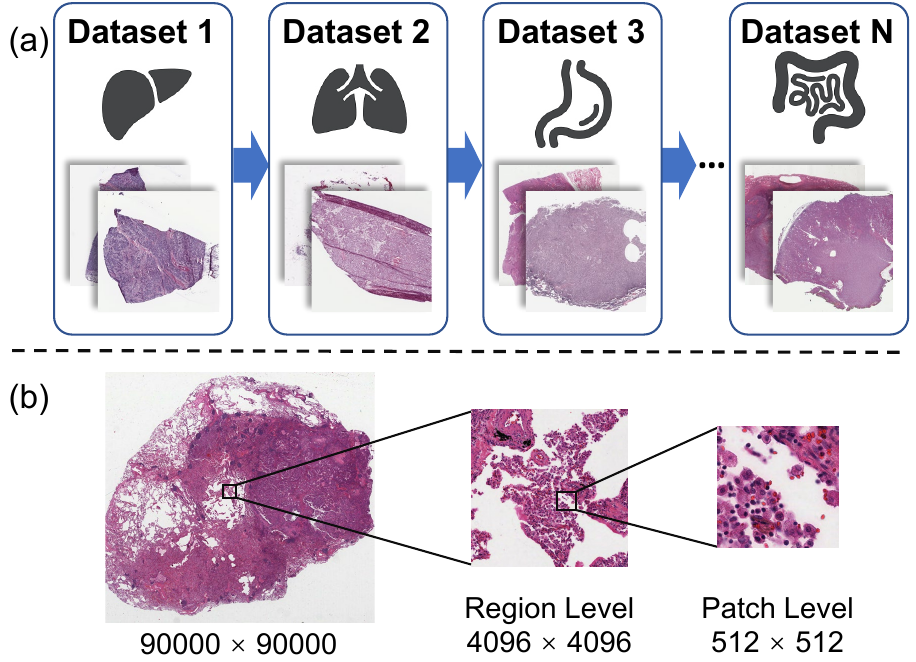} 
\caption{Illustration of concept and challenges of continual WSI analysis. (a) Continual WSI learning aims to alleviate \textit{catastrophic forgetting} while exploiting knowledge from \textit{sequentially incoming} datasets from the different tasks/domains. (b) The hierarchical structure of WSIs raises challenges for network architecture and learning strategy design.}
\label{fig:intro}
% \vspace{-0.4cm}
\end{figure}

Although encouraging results are achieved, these approaches typically adopt the \textit{static model learning} setting~\cite{li2021dual,shao2021transmil,zhang2022dtfd,chen2022scaling,zheng2022kernel,guan2022node}.
However, the WSI imaging technology and staining protocols are not static~\cite{9947057}, which constrains the model performance in new data, and they are subject to an ever-evolving environment in which WSI analysis methods have to adapt in order to remain effectiveness~\cite{lee2020clinical,perkonigg2021dynamic,van2021deep,mahdi2022lifelonger,kaustaban2022characterizing}.
%with the insufficient capability of adapting their behaviors to new tasks/datasets over time, and thus require re-training when facing new WSI tasks or datasets~\cite{de2021continual}. 
%Considering the difficulty and impracticality of preserving the whole previous dataset due to privacy and storage limitations, a naive solution is to fine-tune the pre-trained model on the newly arrived datasets.
%
Although retrain the model every time new data are available is a intuitive solution, but this process can incur high computational and storage costs, especially for WSIs with huge size, while the number of WSIs for different tasks is growing fast~\cite{bandi2023continual}.
Another candidate solution is to train a new model for each new dataset. However, it may be difficult to collect enough data to train a model for each domain (\eg, different stain of WSIs).
Fine-tuning the pre-trained model on the newly arrived datasets is also a candidate solution.
%However, there are large discrepancies in the imaging protocols and the equipment across different sites~\cite{mahdi2022lifelonger}, and there are also large differences in morphological characteristics of cells or tissues of different tumors. Thus, the sequentially arrived datasets are normally heterogeneous with considerable domain shifts.
% However, during the fine-tuning process, the network is prone to concentrate too much on adapting to the feature domain of the current dataset and disrupting the knowledge acquired from previous tasks/datasets, which leads to bad performance on previous tasks. 
However, during the fine-tuning process, the network is prone to concentrate too much on adapting to the feature domain of the current dataset and disrupting the knowledge acquired from previous datasets, which leads to bad performance on previous datasets.
This phenomenon is referred as \textit{catastrophic forgetting}~\cite{boschini2022transfer,de2021continual,lesort2020continual}.

%With the increasingly common deployment of WSIs analysis models in both research and clinical settings, 
% Although previous deep learning approaches can get impressive results in tumor sub-typing or survival prediction tasks, they are obtained with static models incapable of adapting their behavior over time, 
%
% However, gathering large enough WSIs datasets is difficult due to discrepancies in the imaging protocols and the equipment across different site~\cite{mahdi2022lifelonger}.
%
% \ylq{Therefore, a naive solution/common solution is to fine-tune the model trained on one tasks/datasets to another tasks/datas}
% \ylq{Add one sentence like above}
%
% However, in practical application, the former\ylq{former? change another word?} WSIs may become intractable due to privacy issues, or may only be available with a few WSIs due to the limitation of storage space. 
%
% In these cases, the model may suffer from catastrophic forgetting of old concepts as new ones are learned, since the models fit the current input data distribution to the the detriment of previously acquired knowledge~\cite{boschini2022transfer,de2021continual,lesort2020continual}.
% \ylq{Maybe you can refer to https://ieeexplore.ieee.org/document/9908146 to organize the logic of this paragraph.}

\textit{Continual Learning} (CL) was recently proposed~\cite{lopez2017gradient,de2021continual} to overcome the limitations of static model learning and alleviate catastrophic forgetting. 
%
% The aim of CL is to train models with adaptability when meeting new tasks without forgetting the knowledge learned from previous tasks catastrophically, which can make deep learning models much more versatile to the constant growth of medical datasets.
The aim of CL is to train models with adaptability when meeting datasets from new tasks without catastrophically forgetting the knowledge learned from previous tasks, which can make deep learning models much more versatile to the constant growth of medical datasets.
%
%CL is widely researched in the field of natural images, either by using parameter regularization~\cite{kirkpatrick2017overcoming,rebuffi2017icarl,kurle2019continual}, knowledge distillation~\cite{li2017learning,fini2022self,boschini2022class}, and appositely designed architectures~\cite{loo2020generalized,pham2021dualnet,douillard2022dytox}, or by storing and replaying previously saved data~\cite{prabhu2020gdumb,buzzega2020dark,cha2021co2l}. 
%
CL has received much attention in recent years, and it can be achieved either by parameter regularization~\cite{kirkpatrick2017overcoming,rebuffi2017icarl,kurle2019continual}, knowledge distillation~\cite{li2017learning,fini2022self,boschini2022class}, apposite architecture design~\cite{loo2020generalized,pham2021dualnet,douillard2022dytox}, or by data rehearsal-based strategies~\cite{prabhu2020gdumb,buzzega2020dark,cha2021co2l}.
Among these strategies, rehearsal-based methods achieved good performance by replaying a subset of the training data stored in a memory buffer.
% \ylq{something like this}
%
However, the characteristics of WSI pose unique challenges for designing continual WSI analysis frameworks and we are not aware of such frameworks in the existing literature.

WSIs are usually stored at different resolutions, resulting in a hierarchical structure with containing different pathological information, as shown in Figure~\ref{fig:intro} (b). 
%
% Images at different resolutions usually contain different histological information. 
%
For example, patch-level images encompass find-grained cells and tumor cellularity information~\cite{GrahamVRATKR19,Pati2020HACTNetAH,DBLP:conf/iccv/AbousamraBAAYGK21}.
Region-level images mainly characterize the tissue information, such as the extent of tumor-immune localization~\cite{abduljabbar2020geospatial,chen2022scaling,brancati2022bracs}, 
%in representing tumor-infiltrating versus tumor-distal lymphocytes
while slide-level images depict the overall intra-tumoral features~\cite{balkwill2012tumor,javed2020cellular,marusyk2012intra}. 
%Recent works also show that hierarchical models can achieve performance improvement in WSI analysis by utilizing its hierarchical characteristic, and organzing these features in an efficient way~\cite{hou2022h2,chen2022scaling,PatiJFFASBFDRBP22}. 
%
Modeling and utilizing this hierarchical characteristic of WSI is critical for accurate WSI analysis~\cite{hou2022h2,PatiJFFASBFDRBP22}, while it is quite challenging to handle this hierarchical structure when designing a continual WSI analysis framework.
Our preliminary experiments revealed that directly adapting the current CL approaches to hierarchical WSI models will lead to drastic knowledge forgetting of previous datasets (see Section~\ref{Comparison Results}).
Moreover, WSIs are gigapixel images with only slide-level labels, which brings storage and computational challenges for rehearsal-based CL, as it is impractical to store and replay the representative WSI in the limited memory buffer.
To tackle these challenges, we develop a novel WSI continual analysis framework, named \textit{ConSlide} to enable progressive update of a hierarchical WSI analysis architecture by sequentially utilizing the heterogeneous WSI datasets.
%store representative regions during the training process of current task, and replay the augmented WSIs which are reorganized by the saved regions. In this way, the knowledge learned from previous tasks can be preserved, while the model being updated through an asynchronous mechanism.
%
To achieve that, we store a representative region set of past datasets, and then regularly reorganize and replay them during the current model update with an asynchronous updating mechanism.
Specifically, we first design a novel \textit{Hierarchical Interaction Transformer} (HIT) as the backbone to efficiently model and utilize the hierarchical characteristic of WSI. 
HIT is possible to aggregate both fine-grained and coarse-grained WSI features for more comprehensive WSI analysis via its bidirectional interaction within the hierarchical structure. 
Further, to enable the continual update of the designed hierarchical architecture, we follow the rehearsal strategy but develop a novel \textit{Breakup-Reorganize} (BuRo) rehearsal module to tackle the unique challenges of WSI data replay.
Particularly, the BuRo rehearsal module utilizes a random sampling strategy to select and store WSI regions of old tasks in an efficient manner, and then reorganize augmented WSIs of old tasks to improve the knowledge fidelity of old tasks in the replay step.
Based on the augmented old task WSI data, we devise a new asynchronous updating mechanism with the inspiration of \textit{Complementary Learning System (CLS)} theory~\cite{kumaran2016learning}, to encourage the patch-level and region-level blocks to learn generic and task-specific knowledge, respectively, by conducting a nested \textit{Cross-Scale Similarity Learning} (CSSL) task from both old and current WSI data.

With the above careful designs, our framework can preserve the knowledge of previous WSI datasets to mitigate catastrophic forgetting and improve the generalization of the WSI model for more accurate analysis. 
We evaluated our framework on four public WSI datasets from TCGA projects.
The extensive ablation analysis shows the effectiveness of each proposed module. 
Importantly, our \textit{Conslide} achieves more improvement than other compared methods and better trade-off of the overall performance and forgetting on previous datasets under a fair WSI CL setting. 
Our code is available at \href{https://github.com/HKU-MedAI/ConSlide}{https://github.com/HKU-MedAI/ConSlide}.
\begin{figure*}[t]
\centering
\includegraphics[width=1\textwidth]{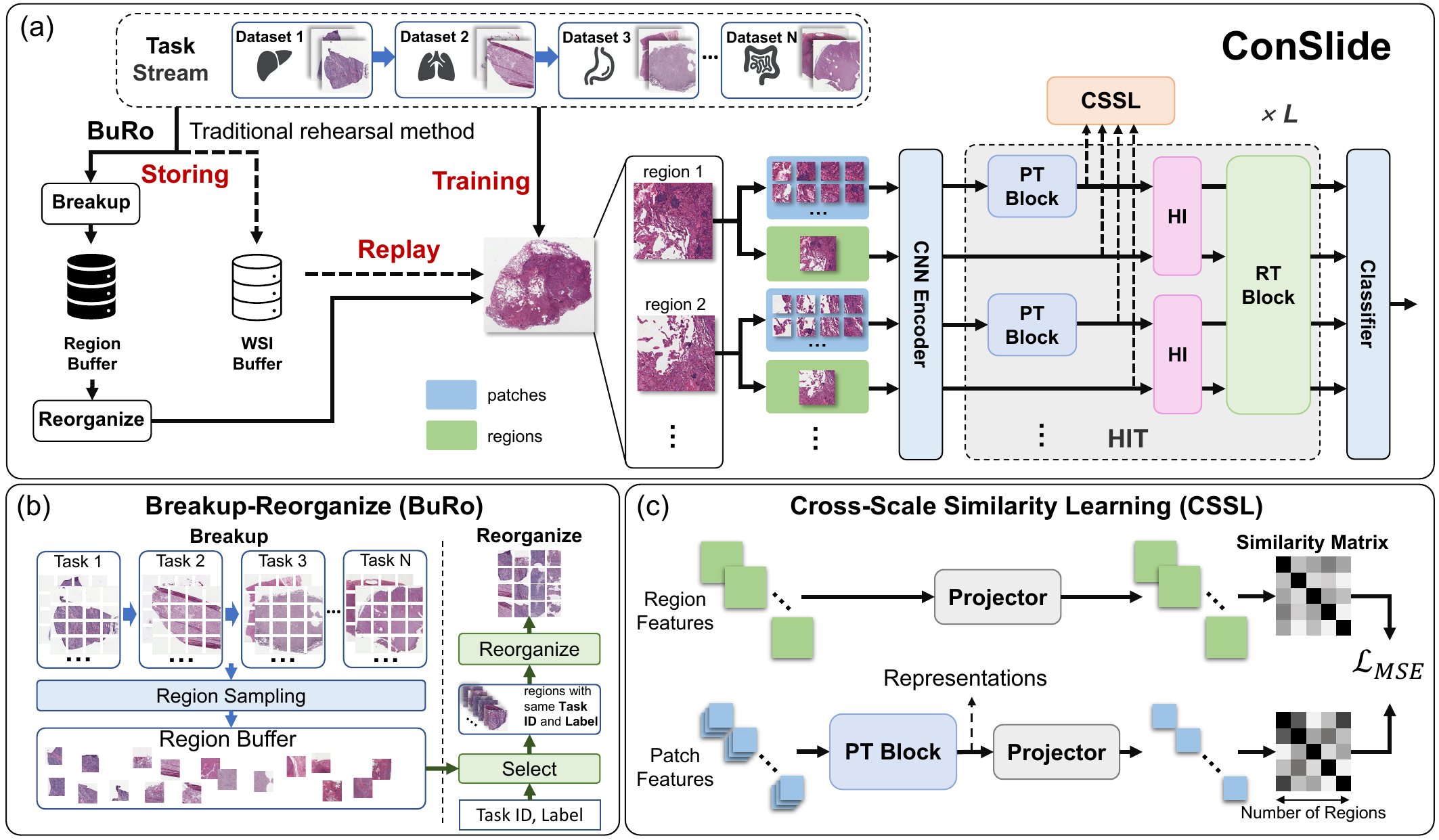} 
% Reduce the figure size so that it is slightly narrower than the column. Don't use precise values for figure width.This setup will avoid overfull boxes.
\caption{Illustration of our proposed continual WSI analysis framework. (a) The architecture and whole process of our proposed \textit{ConSlide} framework, which includes \textit{Training}, \textit{Storing}, and \textit{Replay}. (b) The details of our proposed Breakup-Reorganize (BuRo) rehearsal method. (c) Illustration of the proposed Cross-Scale Similarity Learning (CSSL) module.}
\label{fig:model}
% \vspace{-0.3cm}
\end{figure*}

\section{Related Work}

\para{Multiple Instance Learning for WSI Analysis.}
Multiple Instance Learning (MIL) is widely adopted for WSI analysis.
The conventional MIL approach considers handcrafted aggregators, such as mean-pooling or max-pooling. 
%Due to the sensitivity to singular values, these handcrafted aggregators are easily to be affected by some strange features. 
Recently, attention-based aggregators have shown great potential.
For example, Ilse \textit{et al.}~\cite{ilse2018attention} proposed an attention-based MIL model and obtained the bag embedding according to the contribution of each instance.
Campanella \textit{et al.}~\cite{campanella2019clinical} further extended this strategy to WSI analysis. 
Lu \textit{et al.}~\cite{lu2021data} present an attention-based global pooling operator for weakly-supervised WSI classification by using ResNet-50 as the instance-level feature extraction. 
%
%However, the performance of attention-based aggregators may be affected by some instances that are irrelevant to the classification task. 
Moreover, Li \textit{et al.}~\cite{li2021dual} extend the attention-based aggregators for WSI analysis by combining the max pooling operator and attention-based aggregators with a simple summation.
%, and proved that the handcrafted aggregators are also important for MIL in WSIs analysis.

\para{Hierarchical Structure in WSI.}
To utilize the pyramid structure information of WSI, Li \textit{et al.}~\cite{li2021dual} proposed a multi-scale MIL model that extracts patches at two different resolutions and concatenates features from different resolutions in a pyramidal way. 
Hou \textit{et al.}~\cite{hou2022h2} proposed a hierarchical graph neural network to model the pyramid structure of WSI. 
Recently, Chen \textit{et al.}~\cite{chen2022scaling} proposed a hierarchical Transformer (\ie, HIPT) to leverage the natural hierarchical structure in WSI. 
However, HIPT uses fine-grained features to get the coarse-grained features, and there is no interaction between them.
% \ylq{However, the different level features in HIPT are trained separately and there is no interaction}. Done
%
Instead, our proposed HIT model implements a novel bidirectional information interaction between patch- and region-level features so that it can model WSI in a more effective and accurate manner.
%Although achieving promising performance, Transformer-based models usually suffer from low computation efficiency, and the needs for pre-training and parameter freezing may significantly limit their deployment in end-to-end training scenarios. 
%%%%%%
% Recent research such as H$^2$-MIL~\cite{hou2022h2} propose to regard the WSI pyramid structure as a heterogeneous graph with the "resolution" attribute to explicitly model the features and spatial-scaling relationships of multi-resolution patches. Since the heterogeneous graph is constructed by the spatial relation between patches, it usually lacks the insights in analyzing global information interactions. HIPT~\cite{chen2022scaling} is another promising framework for the analysis of WSIs, which leverages the natural hierarchical structure inherent in WSIs using three levels of Transformer structure~\cite{vaswani2017attention}. Nevertheless, the HIPT largely relies on pre-training and it neglect the coarse-grained features that obtained from region-level images directly.

\para{Continual Learning for Medical Image Analysis.}
Recently, continual learning has been studied in machine learning and computer vision community.
There are different scenarios for continual learning, \eg, class-incremental learning, task-incremental learning, and domain-incremental learning. 
The continual learning strategies can be divided into several categories: regularization-based~\cite{aljundi2018memory,kirkpatrick2017overcoming,li2017learning,zenke2017continual,rebuffi2017icarl,kurle2019continual}, distillation-based~\cite{li2017learning,buzzega2020dark,fini2022self,boschini2022class}, 
architecture design-based~\cite{li2019learn,loo2020generalized,wortsman2020supermasks,pham2021dualnet,douillard2022dytox}, 
and rehearsal-based~\cite{chaudhry2019tiny,prabhu2020gdumb,wu2019large,buzzega2020dark,cha2021co2l}.%
Readers can refer to~\cite{de2021continual} for a detailed survey.
%
% Recently, prompt-based methods~\cite{wang2022learning,wang2022dualprompt} have attracted increasing attention, while the performance of prompt-based methods rely on the well pre-trained Transformer encoder, which constrains its application in continual WSI analysis.
%
Recently, there have been some works studying the CL setting in medical image analysis, such as MRI detection and classification~\cite{baweja2018towards,perkonigg2021dynamic,ozgun2020importance}, X-ray classification~\cite{lenga2020continual}. 
Particularly, some recent works discuss the benchmark of CL in pathology~\cite{mahdi2022lifelonger,kaustaban2022characterizing,bandi2023continual}. 
However, these works are limited in patch-level WSI analysis due to obstacles brought by the gigapixel nature of WSI.
As far as we know, this work is the first one to discuss the challenges and solutions of continual slide-level pathology analysis.
\section{Problem Formulation}

\para{WSI Preprocessing.}
WSI can be formed in a hierarchical structure, \eg, WSI-, region-, and patch-level, respectively.
%
% For an original WSI, we split it into $M$ non-overlapping regions with resolution of $4,096\times 4,096$, and each region is further splitted into $N$ non-overlapping patches with resolution of $512\times 512$. 
For an original WSI $\mathcal{I}$, we split it into $M$ non-overlapping regions $\mathcal{R}$ with size of $4,096\times 4,096$, and each region is further split into $N$ non-overlapping patches $\mathcal{P}$ with size of $512\times 512$, where $\mathcal{I} = [\mathcal{R}_{1}, \mathcal{R}_{2},  \cdots, \mathcal{R}_{M}]$, and $\mathcal{R}_{i} = [\mathcal{P}_{i1}, \mathcal{P}_{i2},  \cdots, \mathcal{P}_{iN}], 1 \leq i \leq M$. 
The regions $\mathcal{R}$ are stored in a downsampling version in which we re-scale the size from $4,096 \times 4,096$ to $512 \times 512$ for downstream processing.
%
% To ease the storage and training, we extract the \textit{patch-level} features $\boldsymbol{P}$ and \textit{region-level} features $\boldsymbol{R}$ with a pre-trained CNN encoder, \eg, ConvNeXt~\cite{liu2022convnet}.
% Therefore, we can obtain the WSI features with the following format $\mathcal{I} = [\boldsymbol{R}_1, \boldsymbol{R}_2,  \cdots, \boldsymbol{R}_M]$, $\mathcal{R}_{i} = [\boldsymbol{P}_{i1}, \boldsymbol{P}_{i2},  \cdots, \boldsymbol{P}_{iN}]$, and $ \boldsymbol{P}_{ij}, \boldsymbol{R}_{i} \in \mathbb{R}^{1 \times C}$, where $C$ means feature channels.
To ease the training, we extract the \textit{patch-level} features $\boldsymbol{P}$ from $\mathcal{P}$ and the \textit{region-level} features $\boldsymbol{R}$ from $\mathcal{R}$ with a pre-trained feature extractor, \eg, ConvNeXt~\cite{liu2022convnet}.
Therefore, we can obtain the region-level and patch-level features with the following format $\boldsymbol{R} = [\boldsymbol{R}_1, \boldsymbol{R}_2,  \cdots, \boldsymbol{R}_M]$, $\boldsymbol{P}_{i} = [\boldsymbol{P}_{i1}, \boldsymbol{P}_{i2},  \cdots, \boldsymbol{P}_{iN}]$, and $ \boldsymbol{P}_{ij}, \boldsymbol{R}_{i} \in \mathbb{R}^{1 \times C}$, where $C$ means feature channels.
% 

% 
% For the ease of storage and training, we use a pre-trained CNN encoder, \eg, ConvNeXt~\cite{liu2022convnet}, to extract \textit{patch-level} features $\mathcal{P}=[\boldsymbol{P}_1, \boldsymbol{P}_2, \cdots, \boldsymbol{P}_M]$, where $\boldsymbol{P}_i=\{\boldsymbol{p}_{i1}, \boldsymbol{p}_{i2}, 
% \cdots, \boldsymbol{p}_{iN}\}, \quad \boldsymbol{p}_{ij} \in \mathbb{R}^{1 \times C}, 1 \leq i \leq M$, where $C$ means feature channels. To obtain additional region-level features, the regions are rescaled $4096\times 4096 \rightarrow  512\times 512$, and \textit{region-level} features $\mathcal{R}=\{\boldsymbol{R}_1, \boldsymbol{R}_2, \cdots , \boldsymbol{R}_M\}, \boldsymbol{R}_i \in \mathbb{R}^{1 \times C}$ are extracted by using the same pre-trained CNN encoder.

\para{Continual Learning for WSI Analysis.}
CL for WSIs is defined as training a model on non-stationary WSI datasets with a series of tasks/classes/domains.
%
%\TODO{For generality, we also different \textit{Dataset} to denote the different classes/tasks/domains.}
% We define a sequence of tasks $\mathcal{D}=\left\{\mathcal{D}_1, \mathcal{D}_2, \cdots, \mathcal{D}_T\right\}$, where the $t$-th task $\mathcal{D}_t=\{\boldsymbol{x}_k, t_k, y_k\}_k$ contains tuples of the WSI sample $\boldsymbol{x}_k$, its corresponding label $y_k$ and optional task identifier $t_k$. In Class-IL scenario, the model needs to predict the classification label $y_k$ when WSI $\boldsymbol{x}_k$ as the input, while in Task-IL scenario, the inputs are the combination of WSI $\boldsymbol{x}_k$ and its corresponding task identifier $t_k$.
We define a sequence of datasets $\mathcal{D}=\left\{\mathcal{D}_1, \mathcal{D}_2, \cdots, \mathcal{D}_T\right\}$, where the $t$-th dataset $\mathcal{D}_t=\{\boldsymbol{x}, y, t\}_k$ contains tuples of the WSI sample $\boldsymbol{x}$, its corresponding label $y$, and optional datasets identifier $t$. The subscript $k$ means there are $k$ samples for dataset $\mathcal{D}_t$. In the class-incremental scenario, the model needs to predict the classification label $y$ with WSI $\boldsymbol{x}$ as the input, while in the task-incremental scenario, the inputs are the combination of WSI $\boldsymbol{x}$ and its corresponding task identifier $t$.

\section{Method}

Our rehearsal-based continual WSI analysis framework is shown in Figure~\ref{fig:model} (a), which includes three steps: \textbf{Training}, \textbf{Storing}, and \textbf{Replay}. 
1) We train the proposed \textit{Hierarchical Interaction Transformer (HIT)} model with the current dataset (and replay data) in the \textbf{Training} step.
2) We adopt the proposed \textit{Breakup-Reorganize (BuRo)} scheme to store a representative set of WSI regions from current datasets to buffer in the \textbf{Storing} step.
3) We \textbf{Replay} the stored WSI data and employ the proposed \textit{asynchronous updating} paradigm to remind the model of the previous dataset knowledge in the next training step.
The whole pipeline is conducted in a closed-loop manner.

\subsection{Hierarchical Interaction Transformer (HIT)}
% In order to aggregate the features of WSIs with hierarchical structure more effectively, we first proposed 
Different from natural images, WSIs are gigapixel images with a hierarchical structure.
Therefore, we design a powerful Hierarchical Interaction Transformer model to abstract the hierarchical structural information of WSI.
As shown in Figure~\ref{fig:model} (a), the proposed HIT model consists of $L$ layers, where each layer contains three components: Patch-level Transformer (PT) block, Region-level Transformer (RT) block, and Hierarchical Interaction (HI) block.
%
% The PT blocks are used to aggregate the patch-level features and the RT blocks are used to aggregate the regions-level feature, while the HI models are adopted to conduct the interaction between these two level blocks.
The PT and RT blocks are used to aggregate the patch-level and region-level features, while the HI block is employed to conduct the interaction between these two level blocks.
%
%Both PT and RT blocks are one layer canonical Transformer, including Multi-head Self-Attention (MSA), Multi-Layer Perceptron (MLP), and Layer Normalization (LN). 

\para{Patch-level Transformer Block.}
Given the patch-level feature set $\boldsymbol{P}^{l-1}_i$ of the $i$-th region $\mathcal{R}_{i}$, the PT block learns the relationship of different patch features in this region and then calculates the output feature
\begin{equation}
\hat{\boldsymbol{P}}^{l}_i = {\rm PT}\left(\boldsymbol{P}^{l-1}_i\right),
\end{equation}
where $l = 1, 2, ..., L$ is the index of the HIT block.
The dimensionality of $\hat{\boldsymbol{P}}^{l}_i$ is $[B, M, N, C]$, where $B$ is the batch size, $M$ is the number of regions, $N$ is the number of patches in on region, and $C$ is the feature dimension.
% , and $L$ is the total number of stacked HIT blocks. 
%The input of first block $\boldsymbol{P}^{0}_i$ is just $\boldsymbol{P}_i$. 
%
${\rm PT}(\cdot)$ is one layer canonical Transformer~\cite{dosovitskiy2020image}, including Multi-head Self-Attention (MSA), Multi-Layer Perceptron (MLP), and Layer Normalization (LN).
The PT blocks share the same parameters among all patches in the same layer. 

\if 0
and the details are as follows.
\begin{equation}
\begin{split}
    \boldsymbol{P}^{l\prime\prime}_i &= \boldsymbol{P}^{l-1}_i + {\rm MSA}\left({\rm LN} \left(\boldsymbol{P}^{l-1}_i \right) \right), \\
    \boldsymbol{P}^{l\prime}_i &= \boldsymbol{P}^{l\prime\prime}_i + {\rm MLP}\left({\rm LN} \left( \boldsymbol{P}^{l\prime\prime}_i \right) \right),
\end{split}
\end{equation}
where $l = 1, 2, ..., L$ is the index of HIT block, and $L$ is the total number of stacked HIT blocks. The input of first block $\boldsymbol{P}^{0}_i$ is just $\boldsymbol{P}_i$. 
%
Then the PT output features $\boldsymbol{P}^{l\prime}_i$ and its corresponding region-level feature $\boldsymbol{R}_i$ are input to HI module.
\fi 

\para{Hierarchical Interaction Block.}
% After conducting the self-attention mechanism among patch-level features,
We design a HI block to fuse the patch-level features and coarse region-level features of the same region in a bidirectional manner. 
Specifically, for region $\mathcal{R}_{i}$ in the $l$-th HIT block, we have its region-level feature $\boldsymbol{R}^{l-1}_i$ and the output feature of the PT block $\hat{\boldsymbol{P}}^{l}_i$ in this layer. 
To get the interacted patch-level feature $\boldsymbol{P}^{l}_i$, we add $\boldsymbol{R}^{l-1}_i$ to $\hat{\boldsymbol{P}}^{l}_i$.
To get the interacted region-level feature $\hat{\boldsymbol{R}}^{l}_i$, we first process the patch-level features $\hat{\boldsymbol{P}}^{l}_i$ with a convolution and a max pooling operator and then add the processed patch-level feature to the region-level feature $\boldsymbol{R}^{l-1}_i$.
The procedure can be represented as
\begin{equation}
\begin{split}
    \boldsymbol{P}^{l}_i &= \hat{\boldsymbol{P}}^{l}_i + \boldsymbol{R}^{l-1}_i, \\
    \hat{\boldsymbol{R}}^{l}_i &= \boldsymbol{R}^{l-1}_i + {\rm Max}\left({\rm Conv}\left(\hat{\boldsymbol{P}}^{l}_i\right)\right).
\label{equ2}
\end{split}
\end{equation}
Since $\hat{\boldsymbol{P}}^{l}_i$ is a set of patches feature vector, we add $\boldsymbol{R}^{l-1}_i$  vector to each vector of set $\hat{\boldsymbol{P}}^{l}_i$.
The max pooling operation in equation~\ref{equ2} is element-wise.
Similar to the PT block, the HI blocks also share the same parameters among all patches in the same layer. 

\para{Region-level Transformer Block.}
% The final process in our HIT model is to aggregate region-level features, which are the output of HI module. The RT block operates on $\mathcal{R}^{l\prime} = [\mathcal{R}^{l\prime}_1, \mathcal{R}^{l\prime}_2, \cdots, \mathcal{R}^{l\prime}_M]$, integrating fused region-level features of $M$ regions. 
The final process in our HIT model is to learn the relationship of different interacted region-level features generated from HI blocks. 
Specifically, the RT block handles the interacted region-level feature set of $M$ regions $\hat{\boldsymbol{R}}^{l} = [\hat{\boldsymbol{R}}^{l}_1, \hat{\boldsymbol{R}}^{l}_2, \cdots, \hat{\boldsymbol{R}}^{l}_M]$, and calculates the output feature 
\begin{equation}
    \boldsymbol{R}^{l}_i = {\rm RT}\left(\hat{\boldsymbol{R}}^{l}_i\right).
\end{equation}
Similar to the PT block, ${\rm RT}(\cdot)$ is also one layer canonical Transformer. 
% integrating region-level features of $M$ regions output by HI module. 
We take the first token of the RT block output at the last layer as the slide-level representation, and then use a linear layer to generate the final slide-level prediction.

\if 0
\begin{equation}
    \boldsymbol{R}^{l}_i = {\rm RT}\left(\boldsymbol{R}^{l\prime}_i\right),
\end{equation}
% and the details is introduced as followed:
\begin{equation}
\begin{split}
    \boldsymbol{R}^{l\prime\prime}_i &= \boldsymbol{R}^{l\prime}_i + {\rm MSA}\left({\rm LN} \left( \boldsymbol{R}^{l\prime}_i \right) \right),\\
    \boldsymbol{R}^{l}_i &= \boldsymbol{R}^{l\prime\prime}_i + {\rm MLP}\left({\rm LN} \left( \boldsymbol{R}^{l\prime\prime}_i \right) \right).
\end{split}
\end{equation}
Similar to the PT block, the RT blocks also share the same parameters among the $L$ layers. 
\fi 

% \faIcon{bullhorn} 
% \para{Discussion.}
% We take the first token of the RT block output at last layer, and use a linear layer to generate the final prediction.
% \TODO{describe how to acquire the final prediction.}
%
\noindent\faIcon{bullhorn} \textbf{Discussion.}
Our proposed HIT model is different from previous hierarchical transformer~\cite{chen2022scaling}. 
We design a novel bidirectional information interaction block between the patch- and region-level features so that our network can model WSI in a more effective and accurate manner.
% \begin{figure}[t]
% \centering
% \includegraphics[width=8cm]{cvpr2023/latex/figures/buffer_method.pdf} 
% % Reduce the figure size so that it is slightly narrower than the column. Don't use precise values for figure width.This setup will avoid overfull boxes.
% \caption{Illustration of Breakup \& Reorganize (BuRo) workflow.}
% \label{fig:buffer}
% \end{figure}

\subsection{Breakup-Reorganize Rehearsal Method}

% What's more, current rehearsal-based CL methods mainly focus on natural images, and they save the whole image to the buffer and replay them without augmentation.
% Existing rehearsal-based CL methods mainly focus on natural images by saving the whole image to the buffer and replaying them without augmentation, which are not suitable for gigapixel WSIs.
Existing rehearsal-based CL methods usually save the whole natural images to the buffer, which is not suitable for gigapixel WSIs.
%
% However, we argue that this paradigm is not suitable for gigapixel WSIs, due to the privacy issue and the storage space restriction. 
%
% Therefore, we introduced a novel Breakup \& Reorganize (BuRo) paradigm, which breakups WSIs and stores samples in region unit. 
% 
% 
% Traditional rehearsal-based CL methods save the whole images to the buffer and replay with the original saved whole images.
% However, with consideration of privacy issue and storage space constrains, this paradigm is not suitable for WSI data.
%
As WSI can be split into non-overlapping regions, we propose a novel rehearsal method, named \textit{Breakup-Reorganize} (RuRo), as shown in Figure~\ref{fig:model} (b) and Algorithm~\ref{alg1}. 
% The overview of BuRo is shown in Figure~\ref{fig:buffer}, and the detailed process is demonstrated in Algorithm~\ref{alg1}.

% \para{Breakup.} For each WSI, we first sample some regions with a certain sampling strategy (\eg, random sampling or sample regions with high/low attention scores), and store these regions in the buffer. 
\para{Breakup.} For each WSI, we first sample some regions randomly, and store these regions in the buffer.
We 
% shall clarify 
highlight that, instead of the original image, the buffer stores region-level features $\boldsymbol{R}_i$ and their corresponding patch-level features $\boldsymbol{P}_i$.

\para{Reorganize.}
During the replaying step, according to a given pair of task id and label each time, we will randomly sample $n$ regions which have the same task id and label from the reservoir $\mathcal{R}^{'}$, and regard these regions as a new augmented WSI $x^{'}$ (since we don't use position information of regions). 
Then we reorganize the selected regions to form a new augmented WSI.
% \faIcon[regular]{star}

\noindent\faIcon{bullhorn} \textbf{Discussion.} We argue that our proposed BuRo method has three advantages.
1) BuRo is resource efficient in continual WSI analysis scenarios, especially when buffer size is limited, as it can store regions from a bigger number of WSI sources;
2) BuRo can boost the diversity of the buffered data by preserving diverse knowledge from previous tasks;
and 3) \textit{Reorganize} operation promotes the generalization ability of the model, as it generates new WSIs with a large number of potential combinations, which means that BuRO is able to ``increase" the sample size without any extra buffer size.
% \textit{Reorganize} operation is similar to mixup~\cite{DBLP:conf/iclr/ZhangCDL18}, and can be regarded as data augmentation on saved samples, which can further promote the generalization in replay process.
% 
% This indicates that BuRo is much more effective in CL scenario on WSIs especially when buffer size is limited. 
% Since \textit{Breakup} operation represent the distribution of previous datasets more accurately by preserving regions from more WSIs, which will also boost the diversity of the buffered data, and thus preserve more knowledge about previous tasks. 
% And the \textit{Reorganize} operation also promote the generalization ability of the model, as it generates new WSIs in the replay step.  
% As the number of potential combinations is big, BuRO is able to increase the sample size without costing any extra buffer size.
%As size of buffer is limited, BuRO is able to increase the diversity of sample for replay by ``'combining'without costing any extra buffer size. 
%
% (Explaination can be: Regions from the same WSI are likely to be homogenous due to the batch effect in histopathology images. Increase the source of the regions will boost the diversity of the buffered data) 

\begin{algorithm}[!t]
	%\textsl{}\setstretch{1.8}
	\renewcommand{\algorithmicrequire}{\textbf{Initialize:}}
	\renewcommand{\algorithmicensure}{\textbf{Output:}}
	\caption{The process of proposed BuRo Method}
	\label{alg1}
	\begin{algorithmic}[1]
	    \REQUIRE dataset $\mathcal{D}$, model $f$, loss function $\ell$, scalar $\alpha$
		\STATE $\mathcal{M} \gets \{\}$
		\FOR {$\mathcal{D}_t$ in $\mathcal{D}$}
            \FOR {$(x_k, t_k, y_k)$ in $\mathcal{D}_t$}
                \STATE  $(\mathcal{R}^{'}, t^{'}, y^{'}) \gets Select(\mathcal{M})$
                \STATE $x^{'} \gets Reorganize(\mathcal{R}^{'})$
                \STATE $loss = \ell(y_k, f(x_k)) + \alpha\ell(y^{'}, x^{'})$
                \STATE $\mathcal{R}^{''} = Sample(Breakup(x_k))$
                \STATE $\mathcal{M} \gets Reservoir(\mathcal{R}^{''}, t_k, y_k)$
            \ENDFOR
        \ENDFOR
% 		\ENSURE  decomposed modes $ \left\{ {{s_k}\left( t \right)} \right\}$, $\left\{ {{\omega _k}\left( t \right)} \right\}$
	\end{algorithmic} 
\end{algorithm}
% \vspace{-0.3cm}
% \subsection{Cross-Scale Similarity Learning for Asynchronous Updating}
\subsection{Asynchronous Updating with Cross-Scale Similarity Learning}

Inspired by CLS theory and other CLS-inspired continual learning models~\cite{parisi2019continual,kemker2018fearnet,9349197,pham2021dualnet}, a continual learning system can be composed of a slow-learner and a fast-learner to learn generic and task-specific features, respectively.
%
%Combining this inspiration and our proposed HIT architecture, the PT blocks should update \textit{slowly} and be mainly used to extract low-level generic features, while the RT blocks need to learn \textit{quickly} on new task and extract high-level task-specific discriminative features.
%In the context of our proposed HIT architecture, we have leveraged this approach to encourage the PT blocks to update gradually while extracting low-level, generic features. Meanwhile, the RT blocks are optimized to learn rapidly and focus on extracting high-level, task-specific discriminative features. This strategy provides a powerful framework for enabling continual learning systems to develop increasingly sophisticated feature extraction capabilities.
In the context of our proposed HIT architecture, we encourage the PT blocks to update \textit{slowly} and extract low-level generic features, while we encourage the RT blocks to learn \textit{quickly} on new tasks and extract high-level task-specific features.
%
% Benefiting from this architecture design, we further propose a novel hierarchical-based paradigm for WSI continual learning, where we encourage the PT blocks to retain the common task-agnostic knowledge in patch-level and RT blocks to extract task-specific knowledge in the region level.
% During the ideal continual learning process, the PT blocks should be updated \textit{slowly} since most tasks share low-level morphology features, while RT blocks need to be learned \textit{quickly} on the new task to extract the task-specific discriminative features.
% Benefiting from this architecture design, we further propose a novel asynchronous updating scheme.
% hierarchical-based paradigm for WSI continual learning, where we encourage the PT blocks to retain the common task-agnostic knowledge in patch-level and RT blocks to extract task-specific knowledge in the region level.
% 
% \sj{These two sentence look very similar. We need to highlight the different otherwise, the reader will think no difference there.}
%
%However, most current SSL methods are designed for natural images, they learn embeddings which are invariant to distortions~\cite{zbontar2021barlow,bardes2021vicreg}, and are not suitable for WSIs which are gigapixel and have hierarchical structure. 
%
Particularly, we design a Cross-Scale Similarity Learning (CSSL) scheme under self-supervised training to facilitate asynchronous updating of the PT block and RT block parameters.
%facilitate the PT block to extract generic features.

%Besides, we assign a smaller loss weight to the PT block. In this way, the PT block and RT block parameters are encouraged to \textit{update asynchronously}.
% To enable that, we design a new Cross-Scale Similarity Learning (CSSL) scheme and assign smaller loss weight for PT block to encourage \textit{asynchronous updating} of the PT block and RT block parameters.

Formally, given a WSI image, we use a linear projector to project its region-level features into region-level projections, and then calculate the cosine similarity among the $M$ region-level projections to form a similarity matrix $\mathcal{C}^r \in \mathbb{R}^{M\times M}$.
Meanwhile, for the patch-level features of the same WSI image, we take the average of the PT block output features within the same region as the ``second" region-level features.
We then use the same linear projector to project them into projections and calculate the second cosine similarity matrix $\mathcal{C}^p \in \mathbb{R}^{M\times M}$ with acquired ``second" region-level projections. 
The objective of the CSSL is to minimize the Mean Squared Error (MSE) loss between these two similarity matrices from different levels:
\begin{equation}
\mathcal{L}_{CSSL}=\sum_{i}\sum_{j}(\mathcal{C}^r_{ij}-\mathcal{C}^p_{ij})^2.
\end{equation}

During the training phase, we conduct the proposed CSSL on the replayed and current WSI data and utilize this supervision to update the PT block parameters.
Also, we calculate the final slide-level prediction loss on the current WSI data and utilize this supervision to update the network parameters.
In this case, the PT blocks need to simultaneously perform well on replayed data and current data, so that they are encouraged to retain the generic low-level knowledge.
Meanwhile, the RT blocks need to perform well on the current dataset, so that they are encouraged to adapt their parameters to the new task.

% \noindent\faIcon{bullhorn} \textbf{Discussion.} Our asynchronous updating paradigm is also inspired by the classic \textit{Complementary Learning System}.\sj{add citation}
% %
% However, different from previous approaches~\cite{buzzega2020dark, pham2021dualnet}, we take advantage of the natural hierarchical structure to design our asynchronous updating paradigm, which avoids explicitly adopting two steps/models and improves the learning efficiency.

\noindent\faIcon{bullhorn} \textbf{Discussion.} The CSSL module constrains the consistency between region- and patch-level matrices calculated by region- and patch-level features respectively, instead of constraining the consistency between region- and patch-level features themselves. 
Although our asynchronous updating paradigm is inspired by the classic complementary learning system~\cite{buzzega2020dark, pham2021dualnet}, we specifically take advantage of the natural hierarchical structure to design our updating paradigm, which avoids explicitly adopting two steps/models and further improves the learning efficiency.

%The recent advanced CL approaches~\cite{pham2021dualnet,wang2022dualprompt} are inspired by the classic \textit{Complementary Learning System (CLS)} theory~\cite{kumaran2016learning,mcclelland1995there}, which adopts one faster and one slower learners for task-specific and task-agnostic knowledge, respectively.
% which adopts a faster learner to learn task-specific knowledge and a slower learner to learn task-agnostic knowledge.
%
\begin{table}
% \small
  \centering
%   \rowcolors {2}{}{blue!10}
  % \resizebox{0.9\columnwidth}{!}{%
  \resizebox{1\columnwidth}{!}{%
  \begin{tabular}{llc}
    \toprule
    Dataset & Tumor type & Cases \\
    \midrule
    \multirow{2}*{NSCLC}    & Lung adenocarcinoma (LUAD) & 492  \\
                            & Lung squamous cell carcinoma (LUSC) & 466  \\
    \midrule
    \multirow{2}*{BRCA}     & Invasive ductal (IDC) &726    \\
                            & Invasive lobular carcinoma (ILC) & 149   \\
    \midrule
    \multirow{2}*{RCC}      & Clear cell renal cell carcinoma (CCRCC) & 498    \\
                            & Papillary renal cell carcinoma (PRCC) & 289   \\
    \midrule
    \multirow{2}*{ESCA}     & Esophageal adenocarcinoma (ESAD) & 65    \\
                            & Esophageal squamous cell carcinoma (ESCC) & 89   \\
    \bottomrule
  \end{tabular}
  }
  \caption{The statistics of continual WSI analysis benchmark.}
  \label{dataset}
  % \vspace{-0.2cm}
\end{table}

\begin{table}[t]
\small
\centering
% \resizebox{0.9\columnwidth}{!}{%
\resizebox{1\columnwidth}{!}{%
\begin{tabular}{l|c|cc}
    \toprule
    Model & \makecell[c]{Patch\\size} & AUC & ACC \\
    \midrule
    CLAM-SB~\cite{lu2021data}   & \multirow{5}*{256} & 0.972 ± 0.008 & 0.770 ± 0.037 \\
    DS-MIL~\cite{li2021dual}            &  & 0.973 ± 0.005 & 0.765 ± 0.010 \\
    TransMIL~\cite{shao2021transmil}    &  & 0.976 ± 0.008 & 0.783 ± 0.067 \\
    HIPT~\cite{chen2022scaling}         &  & 0.977 ± 0.005 & 0.746 ± 0.055 \\
    \textbf{HIT (Ours)}                        &  & \textbf{0.983 ± 0.004} & \textbf{0.833 ± 0.026} \\
    \midrule
    CLAM-SB~\cite{lu2021data} & \multirow{5}*{512}  & 0.967 ± 0.006 & 0.750 ± 0.042 \\
    DS-MIL~\cite{li2021dual}          &             & 0.967 ± 0.004 & 0.760 ± 0.026 \\
    TransMIL~\cite{shao2021transmil}  &             & 0.974 ± 0.003 & 0.765 ± 0.072 \\
    HIPT~\cite{chen2022scaling}       &             & 0.978 ± 0.005 & 0.766 ± 0.029 \\
    \textbf{HIT (Ours)}                      &             & \textbf{0.984 ± 0.003} & \textbf{0.831 ± 0.018} \\
    \bottomrule
\end{tabular}
}
\caption{Model architecture comparison on merged datasets.}
\label{joint}
% \vspace{-0.2cm}
\end{table}

\begin{table*}[t]
\footnotesize
\centering
% \resizebox{0.9\textwidth}{!}{%
\resizebox{1\textwidth}{!}{%
\begin{tabular}{c|l|c|lll|cc}
    \toprule
    CL Type & Method & Buffer size & \makecell[c]{{AUC} ($\uparrow$)} & \makecell[c]{ACC ($\uparrow$)} & \makecell[c]{Masked ACC ($\uparrow$)} & \makecell[c]{BWT ($\uparrow$)} & \makecell[c]{Forgetting ($\downarrow$)} \\
    \midrule
    \multirow{2}*{Baselines} & \makecell[l]{JointTrain (Upper)} & \multirow{2}*{-} & 0.984 ± 0.003 & 0.831 ± 0.018 & 0.902 ± 0.011 & - & - \\
    & \makecell[l]{Finetune (Lower)} &  & 0.700 ± 0.025 & 0.280 ± 0.043 & 0.733 ± 0.040 & -0.224 ± 0.062 & 0.224 ± 0.062 \\
    \midrule
    \multirow{2}*{\makecell[c]{Regularization\\based}} & LwF~\cite{li2017learning} &  \multirow{2}*{-} & 0.721 ± 0.030 $^{***}$ & 0.225 ± 0.013 $^{***}$ & 0.780 ± 0.031 $^{*}$ & -0.160 ± 0.056 & 0.161 ± 0.055 \\
    & EWC~\cite{kirkpatrick2017overcoming}    & & 0.727 ± 0.007 $^{***}$ & 0.273 ± 0.044 $^{***}$ & 0.724 ± 0.042 $^{**}$ & -0.226 ± 0.055 & 0.226 ± 0.055  \\
    \midrule
    \multirow{18}*{\makecell[c]{Rehearsal\\based}}
    & GDumb~\cite{prabhu2020gdumb}    & \multirow{5}*{5 WSIs}& 0.545 ± 0.034 $^{***}$ & 0.201 ± 0.055 $^{***}$ & 0.563 ± 0.040 $^{***}$ & - & -  \\
    & ER-ACE~\cite{caccia2022new}     & & 0.723 ± 0.035 $^{***}$ & 0.271 ± 0.056 $^{***}$ & 0.680 ± 0.031 $^{***}$ & -0.084 ± 0.020 & 0.103 ± 0.047 \\
    & A-GEM~\cite{chaudhry2018efficient} & & 0.803 ± 0.030 $^{***}$ & 0.337 ± 0.057 $^{***}$ & 0.753 ± 0.058 $^{**}$ & -0.213 ± 0.065 & 0.213 ± 0.065 \\
    & DER++~\cite{buzzega2020dark}    &  & 0.781 ± 0.023 $^{***}$ & 0.339 ± 0.034 $^{***}$ & 0.766 ± 0.045 $^{**}$ & -0.169 ± 0.051 & 0.171 ± 0.051 \\
    & \textbf{ConSlide w/o BuRo} &  & 0.852 ± 0.013 $^{***}$ & 0.413 ± 0.009 $^{***}$ & 0.823 ± 0.016 & -0.126 ± 0.022 & 0.131 ± 0.024 \\
    % \cline{3}
    % \rowcolor{blue!5}
    & \cellcolor{blue!5}\textbf{ConSlide} & \cellcolor{blue!5}\makecell[c]{1100 regions \\ ($\approx$ 5 WSIs)} & \cellcolor{blue!5}\textbf{0.915 ± 0.015} & \cellcolor{blue!5}\textbf{0.553 ± 0.033} & \cellcolor{blue!5}\textbf{0.835 ± 0.032} & \cellcolor{blue!5}-0.066 ± 0.023 & \cellcolor{blue!5}0.069 ± 0.021 \\
    % \midrule
    \cline{2-8}
    & GDumb~\cite{prabhu2020gdumb}    & \multirow{5}*{10 WSIs} & 0.618 ± 0.064 $^{***}$ & 0.188 ± 0.064 $^{***}$ & 0.516 ± 0.069 $^{***}$ & - & -  \\
    & ER-ACE~\cite{caccia2022new}     & & 0.762 ± 0.024 $^{***}$ & 0.360 ± 0.060 $^{***}$ & 0.716 ± 0.054 $^{**}$ & -0.066 ± 0.066 & 0.102 ± 0.075  \\
    & A-GEM~\cite{chaudhry2018efficient} & & 0.828 ± 0.028 $^{***}$ & 0.322 ± 0.047 $^{***}$ & 0.814 ± 0.030 & -0.122 ± 0.014 & 0.122 ± 0.014 \\
    & DER++~\cite{buzzega2020dark}    & & 0.860 ± 0.025 $^{***}$ & 0.440 ± 0.056 $^{***}$ & 0.809 ± 0.048 & -0.112 ± 0.051 & 0.113 ± 0.050 \\
    & \textbf{ConSlide w/o BuRo} &  & 0.869 ± 0.017 $^{***}$ & 0.464 ± 0.034 $^{***}$ & 0.822 ± 0.038 & -0.102 ± 0.018 & 0.104 ± 0.016 \\
    % \rowcolor{blue!5}
    & \cellcolor{blue!5}\textbf{ConSlide}       & \cellcolor{blue!5}\makecell[c]{2200 regions \\ ($\approx$ 10 WSIs)} & \cellcolor{blue!5}\textbf{0.931 ± 0.014} & \cellcolor{blue!5}\textbf{0.594 ± 0.053} & \cellcolor{blue!5}\textbf{0.837 ± 0.034} & \cellcolor{blue!5}-0.092 ± 0.026 & \cellcolor{blue!5}0.094 ± 0.023 \\
    \cline{2-8}
    & GDumb~\cite{prabhu2020gdumb}    & \multirow{5}*{30 WSIs}& 0.661 ± 0.040 $^{***}$ & 0.233 ± 0.062 $^{***}$ & 0.609 ± 0.066 $^{***}$ & - & - \\
    & ER-ACE~\cite{caccia2022new}     & & 0.844 ± 0.032 $^{***}$ & 0.469 ± 0.079 $^{***}$ & 0.756 ± 0.020 $^{*}$ & -0.019 ± 0.014 & 0.044 ± 0.021  \\
    & A-GEM~\cite{chaudhry2018efficient} & & 0.855 ± 0.023 $^{***}$ & 0.353 ± 0.119 $^{***}$ & 0.800 ± 0.050 & -0.140 ± 0.067 & 0.144 ± 0.064 \\
    & DER++~\cite{buzzega2020dark}    &  & 0.900 ± 0.024 $^{**}$ & 0.597 ± 0.065 $^{*}$ & 0.840 ± 0.037 & -0.078 ± 0.020 & 0.082 ± 0.023 \\
    & \textbf{ConSlide w/o BuRo} &  & 0.940 ± 0.011 & \textbf{0.668 ± 0.040} & \textbf{0.866 ± 0.036} & -0.051 ± 0.028 & 0.055 ± 0.024 \\
    % \rowcolor{blue!5}
    & \cellcolor{blue!5}\textbf{ConSlide}       & \cellcolor{blue!5}\makecell[c]{6600 regions \\ ($\approx$ 30 WSIs)} & \cellcolor{blue!5}\textbf{0.943 ± 0.007} & \cellcolor{blue!5}0.659 ± 0.022 & \cellcolor{blue!5}0.861 ± 0.017 & \cellcolor{blue!5}-0.075 ± 0.030 & \cellcolor{blue!5}0.076 ± 0.030 \\
    \bottomrule
\end{tabular}
}
\caption{Comparison results among different continual learning methods. The best performances are highlighted as \textbf{bold}. $^{*}$/$^{**}$/$^{***}$ denote there are significant different (paired t-test \textit{p}-value $<$ 0.05/0.01/0.001) between best performances with other comparisons.}
\label{cl_result}
% \vspace{-0.3cm}
\end{table*}

% \begin{table*}[t]
% \small
% \centering
% \resizebox{0.85\textwidth}{!}{
% \begin{tabular}{l|cc|ccc|cc}
%     \toprule
%     Method  & PT Para. & CSSL & \makecell[c]{AUC ($\uparrow$)} & \makecell[c]{ACC ($\uparrow$)} & \makecell[c]{Masked ACC ($\uparrow$)} & \makecell[c]{BWT ($\uparrow$)} & \makecell[c]{Forgetting  ($\downarrow$)} \\
%     \midrule
%     \multirow{2}*{DER++~\cite{buzzega2020dark} + HIT}  
%     &Freeze &  & 0.934 ± 0.013 & 0.666 ± 0.018 & 0.857 ± 0.034 & -0.053 ± 0.029 & 0.057 ± 0.016 \\
%     &Dynamic & & 0.900 ± 0.024 & 0.597 ± 0.065 & 0.840 ± 0.037 & -0.078 ± 0.020 & 0.082 ± 0.023 \\
%     \midrule
%     \multirow{2}*{ConSlide w/o BuRo}
%     &Freeze & $\checkmark$  & 0.934 ± 0.012 & 0.666 ± 0.037 & 0.860 ± 0.037 & -0.059 ± 0.028 & 0.066 ± 0.025 \\
%     & Dynamic & $\checkmark$ & \textbf{0.940 ± 0.011} & \textbf{0.668 ± 0.040} & \textbf{0.866 ± 0.036} & \textbf{-0.051 ± 0.028} & \textbf{0.055 ± 0.024} \\
%     \bottomrule
% \end{tabular}
% }
% \caption{Ablation study of asynchronous updating mechanism in ConSlide w/o BuRo.}
% \label{tab:abla_asy}
% % \vspace{-0.3cm}
% \end{table*}

\begin{table*}[t]
\small
\centering
% \resizebox{0.85\textwidth}{!}{
\resizebox{1\textwidth}{!}{
\begin{tabular}{l|cc|lll|cc}
    \toprule
    Method  & BuRo & CSSL & \makecell[c]{AUC ($\uparrow$)} & \makecell[c]{ACC ($\uparrow$)} & \makecell[c]{Masked ACC ($\uparrow$)} & \makecell[c]{BWT ($\uparrow$)} & \makecell[c]{Forgetting  ($\downarrow$)} \\
    \midrule
    \multirow{4}*{ConSlide}  
    &  &  & 0.860 ± 0.025 $^{***}$ & 0.440 ± 0.056 $^{***}$ & 0.809 ± 0.048 & -0.112 ± 0.051 & 0.113 ± 0.050 \\
    &  & $\checkmark$ & 0.869 ± 0.017 $^{***}$ & 0.464 ± 0.034 $^{***}$ & 0.822 ± 0.038 & -0.102 ± 0.018 & 0.104 ± 0.016 \\
    & $\checkmark$ &  & 0.903 ± 0.013 $^{**}$ & 0.509 ± 0.034 $^{***}$ & 0.833 ± 0.010 & -0.099 ± 0.021 & 0.101 ± 0.023 \\
    & $\checkmark$ & $\checkmark$ & \textbf{0.931 ± 0.014} & \textbf{0.594 ± 0.053} & \textbf{0.837 ± 0.034} & -0.092 ± 0.026 & 0.094 ± 0.023 \\
    % \midrule
    % \multirow{2}*{ConSlide w/o BuRo}
    % &Freeze & $\checkmark$  & 0.934 ± 0.012 & 0.666 ± 0.037 & 0.860 ± 0.037 & -0.059 ± 0.028 & 0.066 ± 0.025 \\
    % & Dynamic & $\checkmark$ & \textbf{0.940 ± 0.011} & \textbf{0.668 ± 0.040} & \textbf{0.866 ± 0.036} & \textbf{-0.051 ± 0.028} & \textbf{0.055 ± 0.024} \\
    \bottomrule
\end{tabular}
}
\caption{Ablation study of RuRo method and asynchronous updating mechanism in ConSlide when buffer size is 10 WSIs (2200 regions).}
\label{tab:abla_asy}
% \vspace{-0.3cm}
\end{table*}

% \begin{table}[t]
% % \small
% \centering
% \caption{Ablation study about the importance of asynchronous mechanism to ASY-HIT model.}
% \label{tab:abla_asy}
% \begin{tabular}{l|cc}
%     \toprule
%     Metric & HIT & HIT + ASY \\
%     \midrule
%     \textbf{AUC}& 0.900 ± 0.024 & \textbf{0.940 ± 0.011} \\
%     ACC         & 0.597 ± 0.065 & \textbf{0.668 ± 0.040} \\
%     Masked ACC  & 0.840 ± 0.037 & \textbf{0.866 ± 0.036} \\
%     BWT         & -0.078 ± 0.020& \textbf{-0.051 ± 0.028} \\
%     Forgetting  & 0.082 ± 0.023 & \textbf{0.055 ± 0.024} \\
%     \bottomrule
% \end{tabular}
% \end{table}

\section{Experiments}

\subsection{Datasets}
%  We employ the public TCGA datasets to evaluate our framework.
 %The public pathology datasets greatly facilitate the development of WSI analysis. 
 As we are the first to study WSI continual learning framework, we setup a WSI continual learning benchmark as shown in Table~\ref{dataset}.
%  to compare our method with other related approaches.
 %
%  Specifically, the benchmark contains a sequence of WSI-based tumor subtype classification tasks, which is composed of four public WSI datasets from The Cancer Genome Atlas (TCGA) repository: TCGA non-small cell lung carcinoma (NSCLC), TCGA invasive breast carcinoma (BRCA), TCGA renal cell carcinoma (RCC), and TCGA esophageal carcinoma (ESCA). 
%   Specifically, the benchmark
It contains a sequence of WSI-based tumor subtype classification tasks, which includes four public WSI datasets from The Cancer Genome Atlas (TCGA) repository: non-small cell lung carcinoma (NSCLC), invasive breast carcinoma (BRCA), renal cell carcinoma (RCC), and esophageal carcinoma (ESCA). 
Note that we simply choose two subtypes of TCGA-RCC for consistency with other datasets (both two subtypes). 
% 
%  The detailed dataset information is illustrated in Table~\ref{dataset}.
 %
 
\subsection{Experimental Setting}
\label{experimental_setting}
We mainly adopt the class-incremental setting to compare our framework with other methods. 
% 
% We regard the NSCLC, BRCA, RCC, and ESCA datasets as the sequentially arrived datasets/tasks and gradually conduct continual learning.
We define the arrival sequence for four datasets as NSCLC, BRCA, RCC, and ESCA to conduct continual learning.
In our designed benchmark, each dataset contains two categories.
For the model development of the first dataset, it is defined as a two-class learning task. 
While for each new arrival dataset, we increase the class number by two at each time and conduct multi-class learning.
The following results are reported by ten-fold cross-validation.

\para{ Evaluation Metrics.}
We assess the model's final performance with Area Under the Curve (\textbf{AUC}), Accuracy (\textbf{ACC}), and Masked Accuracy (\textbf{Masked ACC}) of all the previous and current datasets after continual learning.
Note that the Masked ACC can reflect the performance of CL on the task-incremental scenario and it is calculated by masking irrelevant categories from different datasets.
Besides the final performance, we assess the overall performance over the entire continual learning time span by \textit{Backward Transfer (BWT)} and \textit{Forgetting} following previous works \cite{boschini2022transfer,fini2022self} for reference only.

% Therefore, following previous works\cite{boschini2022transfer,fini2022self}, we calculate \textit{Backward Transfer (BWT)}, and \textit{Forgetting} for a more comprehensive aseesment.
% to evaluate the overall performance in the entire continual learning time span.
%
% \TODO{The detailed definitions of these metrics can be found in supplementary materials.How about add one citation here. so that we don't need the definition.}
%
% It is worth noting that AUC, ACC, and Masked ACC evaluate the overall performance of our framework, while the BWT and Forgetting cannot demonstrate the model's absolute performance on single task and often serve as reference metrics~\cite{boschini2022transfer,fini2022self}.

\subsection{Implementation Details}

Following the WSI pre-processing in~\cite{lu2021data},
we first segment the foreground tissue and then crop $4,096 \times 4,096$ tiles from the segmented tissue at 20X magnification as region-level images.
Each cropped region is further divided into 64 non-overlapping $512 \times 512$ patches as the patch-level images. 
We adopted ConvNeXt~\cite{liu2022convnet} as the pre-trained feature extractor to extract both the patch-level features and region-level features from the corresponding images. 
For a fair comparison, we adopt the same patch-level and region-level features as the model input for all other methods.

\subsection{Experimental Results}
\label{Comparison Results}

\para{Comparison with Other WSI Analysis Approaches.}
We first compare our proposed HIT backbone with several state-of-the-art WSI analysis approaches to show the effectiveness of utilizing hierarchical structure information for WSI analysis:
1) single-attention-branch CLAM-SB~\cite{lu2021data}, 
2) non-local attention pooling-based  DS-MIL~\cite{li2021dual}, 
3) correlated MIL-based TransMIL~\cite{shao2021transmil}, and 4) a recent hierarchical structure-based HIPT~\cite{chen2022scaling}.
%
% For this part, we merge the mentioned four TCGA datasets together and formulate a multi-class classification task (\ie, eight-class) to predict the tumor type of each WSI by taking $256\times 256$ and $512\times 512$ patch as inputs. 
We merge the mentioned four TCGA datasets together and formulate an eight-class classification task to predict the tumor type of each WSI.
% by taking $256\times 256$ and $512\times 512$ patches as inputs. \sj{I put the patch size in the next para} 
% \TODO{And more experimental details can be found in supplementary materials.}
% \ylq{need to add more expermental details in supp.}

The average per epoch training time under JointTrain setting with a single RTX 3090 GPU card are 1.32min, 1.97min, 2.87min, 1.72min, and 1.90min for CLAM-SB, DS-MIL, TransMIL, HIPT, and our HIT, respectively. 
Table~\ref{joint} shows the comparison results with the above approaches under two different patch size settings of $256$ and $512$.
It is clear that our proposed HIT achieves the best AUC and ACC score under both settings.
% over other SOTA WSI analysis methods under both two settings.
%
Specifically, HIT outperforms the SOTA approaches with 0.6\% and 5.0\% improvement on AUC and ACC respectively when patch size is $256\times256$, and also acquires a performance improvement of 0.6\% AUC and 6.5\% ACC respectively when patch size is $512\times512$.
HIPT~\cite{chen2022scaling} also utilizes the hierarchical information of WSI for analysis and thus outperforms other WSI analysis methods on the AUC metric.
%
% However, there is no interaction among different level features in HIPT.
%
Benefiting from the proposed bidirectional hierarchical interaction scheme, HIT achieves better results than HIPT where no interaction among different level features is utilized, which further shows the effectiveness of the proposed HIT block.
% to mine hierarchical structure for WSI analysis.  \sj{HIPT also utilize hierarchical information, so we cannot generate this conclusion.}

\para{Comparison with Other Continual Learning Approaches.}
% We then compare our framework with other continual learning strategies on the slide-level WSI analysis task. \sj{same with the subsection title.}
%
\begin{table*}[!t]
\small
\centering
% \resizebox{0.92\textwidth}{!}{
\resizebox{1\textwidth}{!}{
\begin{tabular}{l|c|ccc|cc}
    \toprule
    Method & Buffer size & \makecell[c]{{AUC} ($\uparrow$)} & \makecell[c]{ACC ($\uparrow$)} & \makecell[c]{Masked ACC ($\uparrow$)} & \makecell[c]{BWT ($\uparrow$)} & \makecell[c]{Forgetting ($\downarrow$)} \\
    \midrule
    DER++~\cite{buzzega2020dark}    & \multirow{2}*{5 WSIs} & 0.787 ± 0.017 & 0.349 ± 0.018 & 0.799 ± 0.034 & -0.100 ± 0.022 & 0.109 ± 0.030 \\
    \textbf{ConSlide w/o BuRo}             & & 0.822 ± 0.009 & 0.336 ± 0.022 & 0.805 ± 0.028 & -0.090 ± 0.046 & 0.112 ± 0.049 \\
    \cellcolor{blue!5}\textbf{ConSlide} & \cellcolor{blue!5}\makecell[c]{1100 regions \\ ($\approx$ 5 WSIs)} & \cellcolor{blue!5}\textbf{0.893 ± 0.025} & \cellcolor{blue!5}\textbf{0.426 ± 0.047} & \cellcolor{blue!5}\textbf{0.839 ± 0.039} & \cellcolor{blue!5}-0.005 ± 0.054 & \cellcolor{blue!5}0.045 ± 0.024 \\
    \midrule
    DER++~\cite{buzzega2020dark}    & \multirow{2}*{10 WSIs} & 0.819 ± 0.015 & 0.390 ± 0.018 & 0.787 ± 0.031 & -0.111 ± 0.046 & 0.120 ± 0.051 \\
    \textbf{ConSlide w/o BuRo}             & & 0.867 ± 0.029 & 0.427 ± 0.017 & \textbf{0.839 ± 0.026} & -0.033 ± 0.040 & 0.052 ± 0.020 \\
    \cellcolor{blue!5}\textbf{ConSlide} & \cellcolor{blue!5}\makecell[c]{2200 regions \\ ($\approx$ 10 WSIs)} & \cellcolor{blue!5}\textbf{0.896 ± 0.026} & \cellcolor{blue!5}\textbf{0.453 ± 0.052} & \cellcolor{blue!5}0.829 ± 0.040 & \cellcolor{blue!5}-0.040 ± 0.055 & \cellcolor{blue!5}0.076 ± 0.035 \\
    \midrule
    DER++~\cite{buzzega2020dark}    & \multirow{2}*{30 WSIs} & 0.883 ± 0.032 & 0.527 ± 0.037 & 0.852 ± 0.027 & -0.036 ± 0.041 & 0.064 ± 0.024 \\
    \textbf{ConSlide w/o BuRo}             & & 0.914 ± 0.004 & \textbf{0.543 ± 0.032} & 0.849 ± 0.036 & -0.045 ± 0.032 & 0.058 ± 0.035 \\
    \cellcolor{blue!5}\textbf{ConSlide} & \cellcolor{blue!5}\makecell[c]{6600 regions \\ ($\approx$ 30 WSIs)} & \cellcolor{blue!5}\textbf{0.918 ± 0.008} & \cellcolor{blue!5}0.499 ± 0.025 & \cellcolor{blue!5}\textbf{0.854 ± 0.039} & \cellcolor{blue!5}-0.021 ± 0.039 & \cellcolor{blue!5}0.058 ± 0.032 \\
    \bottomrule
\end{tabular}
}
\caption{Continual learning results of different methods when the sequence of datasets is \textit{reversed}.}
\label{cl_result_reverseorder} 
% \vspace{-0.3cm}
\end{table*}
Before comparison to continual learning methods, 
we jointly train a model with the merged datasets (JointTrain), which can be regarded as upper-bound, and train these four datasets on-by-one with the fine-tuning scheme (Finetune), which can be regarded as lower-bound. 
We then compare with several state-of-the-art continual learning approaches, including regularization-based methods LwF~\cite{li2017learning} and EWC~\cite{kirkpatrick2017overcoming}, and rehearsal-based methods GDumb~\cite{prabhu2020gdumb}, ER-ACE~\cite{caccia2022new}, A-GEM~\cite{chaudhry2018efficient}, and DER++~\cite{buzzega2020dark}, under different buffer size settings (\ie, 5, 10, and 30 WSIs). 
We re-implement these continual learning approaches with the proposed HIT as the backbone for a fair comparison.
% \TODO{The detailed setting of each method can be found in supplementary materials.}
%For previous CL methods, directly applying them on WSI data is not feasible, since most of them are based on CNN encoder, like ResNet~\cite{he2016deep}. Thus, we reproduce serveral powerful baselines with the proposed HIT as backbone to realize a fair comparison. The reproduced approaches include:  We compare our proposed ASY-HIT and BuRo with serveral strong CL baselines, including: regularization-based methods (LwF, EWC) and rehearsal-based methods (DER++, GDumb, ER-ACE), 
% 
% Table~\ref{cl_result} shows the continual learning results of baselines (\ie, JointTrain and Finetune) and various approaches.
Table~\ref{cl_result} shows the comparison results. 
% continual learning results of baselines (\ie, JointTrain and Finetune) and various approaches.

The performance of JointTrain is significantly higher than Finetune (improved 28.3\% in AUC, 52.7\% in ACC, 15.6\% in Masked ACC), confirming that the naive fine-tuning approach suffers from catastrophic forgetting when meeting new WSI datasets. 
Moreover, it is observed that regularization-based methods perform worse than rehearsal-based methods and only achieve a slight improvement compared with the lower-bound baseline in AUC metric, which is consistent with natural image observations~\cite{buzzega2020dark}.
By comparing rehearsal-based methods under different buffer sizes, it is observed that GDumb~\cite{prabhu2020gdumb} performs worse than the others, although GDumb has good performance in the natural image-based continual learning setting.
Particularly, the reason may be that the buffered WSI samples cannot accurately represent the task-specific feature distribution, thus making the model difficult to preserve enough old task knowledge during the replay step.
This phenomenon validates that it is non-trivial to extend the natural image CL approaches to WSI due to the unique challenges of analyzing gigapixel WSI.
The rehearsal-based ER-ACE~\cite{caccia2022new}, A-GEM~\cite{chaudhry2018efficient}, and DER++~\cite{buzzega2020dark} can achieve better performance than GDumb due to advanced learning strategies.
Compared with DER++ under different buffer size situations, the proposed \name w/o BuRo already consistently produces 7.1\%, 0.9\%, and 4.0\% improvements in AUC, 7.4\%, 2.4\%, and 7.1\% improvements in ACC, and 5.7\%, 1.3\%, and 2.6\% improvements in Masked ACC, validating the effectiveness of the proposed hierarchical-based asynchronous updating continual learning paradigm. 
By incorporating the proposed BuRo module, \name can further boost the continual learning performance.
Note that the BWT and Forgetting metrics of our proposed \name are not the best among these methods, since there is a trade-off between absolute performance and forgetting resistance, and these two metrics cannot comprehensively evaluate the effectiveness of our framework.\footnote{If the absolute performance is low, it is easy to maintain the similar performance during the model updating and thus have a large BWT and low Forgetting value.}
Following previous works~\cite{boschini2022transfer,fini2022self}, we list BWT and Forgetting metrics as reference only.
% \ylq{Following previous works~\cite{XXXX}, we list BWT and Forgetting metrics as reference only}.

\if 0
\begin{figure}[t]
\centering
\includegraphics[width=0.8\columnwidth]{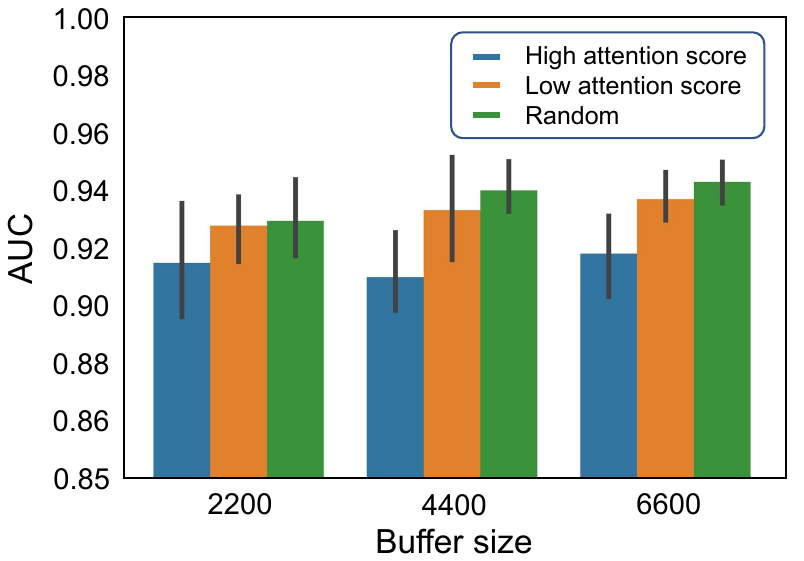} 
% Reduce the figure size so that it is slightly narrower than the column. Don't use precise values for figure width.This setup will avoid overfull boxes.
\caption{Performance of different sampling methods for BuRo module with different buffer sizes.}
\label{fig:sample_method}
\vspace{-0.3cm}
\end{figure}
\fi

\if 0

\begin{table}[htbp]
\small
\centering
\begin{tabular}{ccc|cc}
    \toprule
    PL & RL & HI module & AUC & ACC \\
    \midrule
    $\checkmark$ & & & 0.979 ± 0.003 & 0.805 ± 0.018 \\
    & $\checkmark$ & & 0.977 ± 0.006 & 0.784 ± 0.050 \\
    $\checkmark$ & $\checkmark$ & & 0.981 ± 0.006 & 0.805 ± 0.033 \\
    $\checkmark$ & $\checkmark$ & $\checkmark$ & \textbf{0.983 ± 0.004} & \textbf{0.807 ± 0.037} \\
    \bottomrule
\end{tabular}
\caption{Ablation study on in HIT model.}
\label{tab:abla_hit}
\end{table}

\fi

\subsection{Analysis of Our Framework}

\begin{table}[t]
\small
\centering
\begin{tabular}{ccc|ll}
    \toprule
    PL & RL & HI module & AUC & ACC \\
    \midrule
    $\checkmark$ & & & 0.979 ± 0.003 $^{**}$ & 0.805 ± 0.018 $^{*}$ \\
    & $\checkmark$ & & 0.977 ± 0.006 $^{**}$ & 0.784 ± 0.050 $^{**}$ \\
    $\checkmark$ & $\checkmark$ & & 0.981 ± 0.006 $^{*}$ & 0.805 ± 0.033 $^{*}$ \\
    $\checkmark$ & $\checkmark$ & $\checkmark$ & \textbf{0.984 ± 0.003} & \textbf{0.831 ± 0.018} \\
    \bottomrule
\end{tabular}
\caption{Ablation study on in HIT model. $^{*}$/$^{**}$/$^{***}$ denote there are significant different (paired t-test \textit{p}-value $<$ 0.05/0.01/0.001) between best performances with other comparisons.}
\label{tab:abla_hit}
\end{table}

\para{Additional Ablation Analysis of HIT Model}
We further conduct an ablation study for the proposed HIT module to investigate the effectiveness of each component.
Table~\ref{tab:abla_hit} lists the ablation results.
It is observed that the performance will degrade if we drop either region-level or patch-level features and the performance degradation is large if we drop patch-level features (comparing the first three rows).
By incorporating the hierarchical interaction module, the performance is further boosted (comparing the last two rows).

We first conducted experiments on the HIT model without region-level features, and the performance degrades obviously, while the degradation become larger for HIT model without patch-level features.
This pinpoints that the fine-grained information of patch-level features and the coarse-grained information of region-level features are both vital for WSIs analysis.
After that, we conducted experiments on HIT model without HI module, \ie, replacing the convolution operation and max pooling in HI module with simple mean pooling, and it indicate the the combination of patch- and region-level features can further promote the performance. 
However, such simple combination is still less representative than the proposed HI module, which reveals that the HI module is effective for fusing features of different scales.

% \begin{figure*}[t]
% \centering
% \includegraphics[width=0.75\textwidth]{iccv2023/figures/attention_map.pdf} 
% % Reduce the figure size so that it is slightly narrower than the column. Don't use precise values for figure width.This setup will avoid overfull boxes.
% \caption{The attention heatmaps of four samples from four consecutive tasks in four different continual learning process. The figures with orange/green/blue box are the heatmaps of current/past/unseen tasks.}
% \label{fig:attention}
% \end{figure*}

\begin{figure}[t]
\centering
\includegraphics[width=\columnwidth]{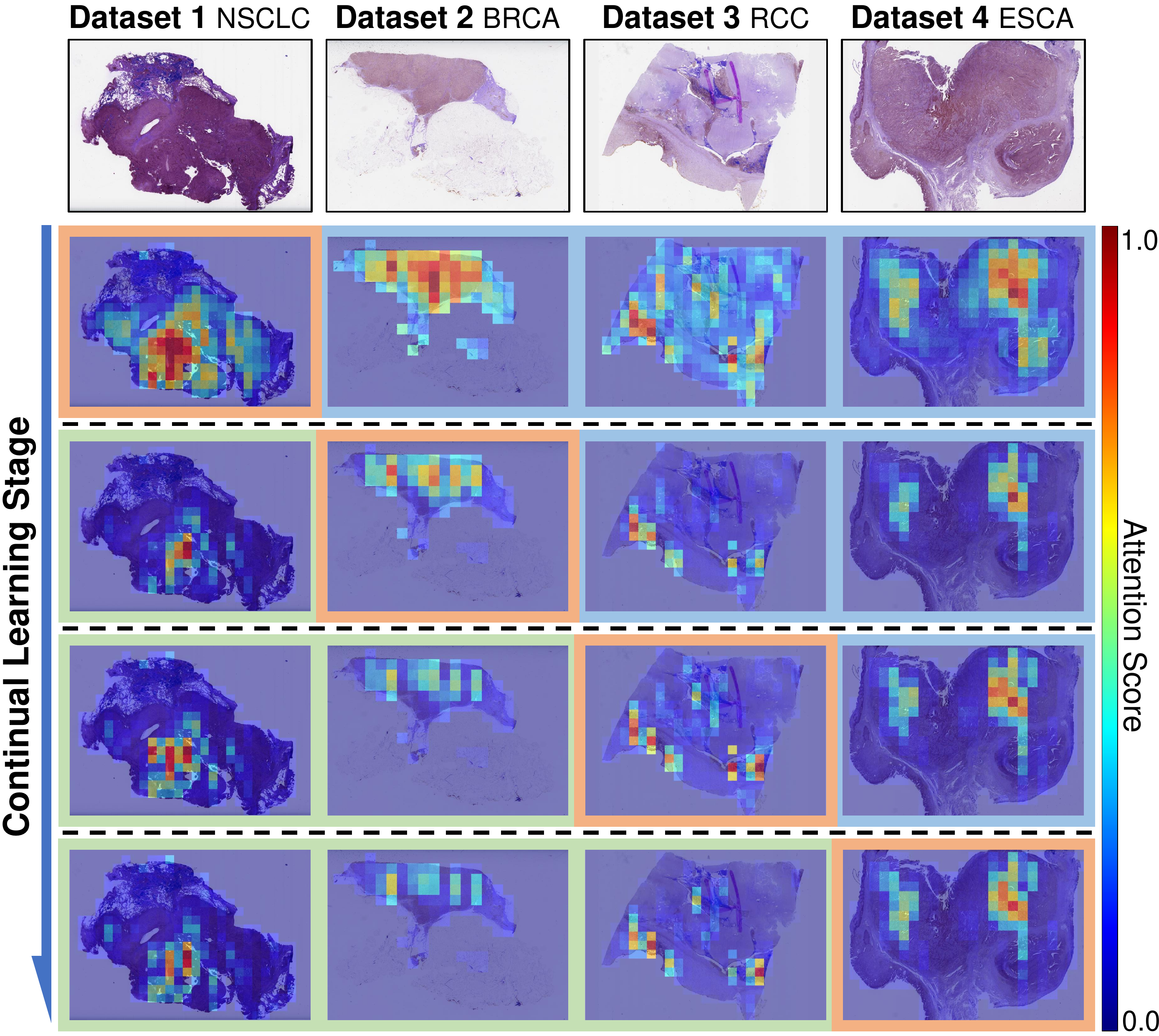} 
% Reduce the figure size so that it is slightly narrower than the column. Don't use precise values for figure width.This setup will avoid overfull boxes.
\caption{The region-level attention maps of four samples from four consecutive tasks in four different continual learning stages. The figures with orange/green/blue box are the attention maps of current/past/unseen tasks.}
\label{fig:attention}
% \vspace{-0.3cm}
\end{figure}

\para{Effectiveness of BuRo Module.}
To figure out the effectiveness of the proposed Breakup-Reorganize rehearsal method and asynchronous updating mechanism, we did more ablation study about the BuRo and CSSL module of \name framework. The results are summarized in Table~\ref{tab:abla_asy}.
The comparison between {\name w/o BuRo} and {\name w/ BuRo} in different buffer size also reported in Table~\ref{cl_result}, which shows the effectiveness of the proposed Breakup-Reorganize Rehearsal method. 
Note that for a fair comparison, we set up the buffer size of BuRo in an approximate way (\ie, using the average region numbers of each WSI).
%
% By storing a representative set of regions not the whole WSIs, BuRo can save all kinds of regions from more WSI sources.
By storing a representative set of regions, not the whole WSIs, BuRo can save diverse regions from more WSI sources with a fixed buffer size.
% 
% In this way, BuRo is able to better preserve the diversity and feature distribution of the original datasets with a fixed buffer size.
By reorganizing the regions to generate more new replay WSI samples, BuRo also promotes the generalization ability of the model.
The performance is thus benefited from the above advantages (\ie, 6.3\%, 6.2\%, 0.3\% AUC improvement under 5, 10, and 30 WSI buffer sizes).
We can observe that the marginal benefit of BuRo is saturated with the increase of buffer size.
Therefore, BuRo is much more effective in CL scenarios on WSIs, especially when buffer size is limited.

\if 0
\para{Analysis of the Sampling Strategy in BuRo Module.}
We further investigate the impact of different region sampling strategies for BuRo.
%
Figure~\ref{fig:sample_method} lists the performance of three different sampling strategies under different buffer sizes.
%
We observed that sampling regions with high attention scores performed worst among the three strategies under all settings.
%
As regions with high attention scores are highly correlated with the discriminative regions in the tumor typing task (\ie, tumor regions), 
%
BuRo will tend to generate new WSIs full of discriminative regions (\ie, tumor regions) for reply training under such a sampling strategy.
%
However, training with such new WSIs will ease the reply training process and thus blur the classification boundary.
%
Therefore, the strategy sampling regions with high attention scores are likely to suffer embarrassing performance when meeting real WSIs with only a small number of discriminative regions (\ie, tumor regions) during the inference phase.
%
Another interesting fact is that random sampling performed best consistently under all settings, beating sampling regions with low-attention scores.
%
This is because random sampling can better preserve the diversity and feature distribution of original datasets by preserving all kinds of regions equally.
%
On the contrary, attention-based sampling strategies tend to save homogeneous regions (\ie, regions with close attention scores), which leads to a distorted representation of the original dataset.
%
\fi

\para{Analysis of Asynchronous Updating.}
% To figure out the effectiveness of the proposed asynchronous updating mechanism, and validate the complex model update for hierarchical WSI models on multiple sequential tasks, we tested DER++ and our \name w/o BuRo with Freeze and Dynamic PT parameters. 
% To figure out the effectiveness of the proposed asynchronous updating mechanism and BuRo rehearsal method, we tested DER++ and our \name w/o BuRo with Freeze and Dynamic PT parameters. 
% %
% The results are summarized in Table~\ref{tab:abla_asy}.
%
The comparison between \name w/o CSSL and \name w/ CSSL in Table~\ref{tab:abla_asy} shows the effectiveness of the proposed asynchronous updating mechanism.
Compared with \name w/o CSSL, our asynchronous updating mechanism based on the CSSL consistently achieved better performance in all metrics no matter whether there is BuRo module, demonstrating the prominence of the asynchronous updating mechanism.
%
% Specifically, for DER++, if we adopt Dynamic PT, the RT will need to adapt to the challenging Synchronous updates from both the dataset and PT parameter, thus leading to unstable optimization. 
%
% Therefore, better performance is obtained by simply freezing the updates from PT.
%
% However, for \name w/o BuRo, with the proposed CSSL method, the PT and RT are forced to update their parameters in an Asynchronous manner. 
%
%In this case, the RT can adapt to updates from both the dataset and PT parameter more easily, and thus achieve the most promising results in the Dynamic end-to-end training process.

\para{Impact of Changing Task Orders.}
We conducted experiments to verify the generalization of our framework. 
By reversing the sequence of the datasets in Section \ref{experimental_setting},
% Table~\ref{cl_result}
we compare the performance of our framework with the strongest CL baseline DER++ and show the results in
Table~\ref{cl_result_reverseorder}.
% summarizes the experimental results. 
We can observe that the proposed \name w/o BuRo still achieved obvious performance increases in AUC, ACC and Masked ACC than DER++ under all settings.
Besides,  \name with BuRo also shows further performance improvement in the reversed datasets, especially under limited buffer size (\ie, 5 WSIs).
These results together demonstrate the robustness of our proposed \name framework.

\para{Attention Map Visualization.}
We further visualized the attention heads of the RT block in the HIT model by using ``Attention Rollout''~\cite{abnar2020quantifying} algorithm, and get the region-level attention maps of four samples from four consecutive tasks in different continual learning stages in Figure~\ref{fig:attention}.
First, we average the attention weights of RT blocks across all heads and then recursively multiplied the weight matrices of all layers to get the attention map between all tokens (including region-level tokens and one class token). After that, we use the $M$ dimension attention score of the class token to represent the attention map with single heatmap colour.
The different stages are separated by dash lines, and the figures with orange/green/blue boxes are the attention maps of current/past/unseen tasks in each continual learning stage.
From the maps with green boxes, it can be observed that the learned importance of tumor regions for the same sample is maintained during different stages, showing that the knowledge of previous tasks can be preserved in the proposed ConSlide framework.
Besides, from the maps with blue boxes, we can find that ConSlide can assign high attention scores to tumor regions even for unseen tasks, indicating that the model trained in previous tasks can provide some prior knowledge for subsequent tasks.

\if 0
\para{Ablation Analysis of HIT Module.}
We conduct an ablation study for the proposed HIT module to investigate the effectiveness of each component.
Table~\ref{tab:abla_hit} lists the ablation results.
%
It is observed that the performance will degrade if we drop either region-level or path-level features and the performance degradation is large if we drop path-level features (comparing the first three rows).
%
By incorporating the hierarchical interaction module, the performance is further boosted (comparing the last two rows).

We first conducted experiments on the HIT model without region-level features, and the performance degrades obviously, while the degradation become larger for HIT model without patch-level features.
%
This pinpoints that the fine-grained information of patch-level features and the coarse-grained information of region-level features are both vital for WSIs analysis.
%
After that, we conducted experiments on HIT model without HI module, \ie, replacing the convolution operation and max pooling in HI module with simple mean pooling, and it indicate the the combination of patch- and region-level features can further promote the performance. 
%
However, such simple combination is still less representative than the proposed HI module, which reveals that the HI module is effective for fusing features of different scales.
\fi 
\section{Conclusion}

In this paper, we propose ConSlide, an asynchronous hierarchical interaction transformer with breakup-reorganize rehearsal for continual WSI analysis.
Our ConSlide sheds light on future WSI-based continual learning, due to its carefully detailed key components to deal with the WSI challenges of catastrophic forgetting, the huge size of the image, and efficient utilization of hierarchical structure.
The proposed BuRo rehearsal module is specifically designed for WSI data replay with efficient region storing buffer and WSI reorganizing operation.
As we adopt a transformer-based backbone and rely on a low inductive bias of spatial structure, BuRo would not significantly influence the capability of ConSlide to preserve the previous task information, although the reorganization operation in BuRo will interrupt the spatial structure of WSI. 
%
%We perform ablation studies on four WSI datasets of different tasks to evaluate our method and compare it with other methods under a fair WSI-based continual learning setting to demonstrate its superiority in the effectiveness and better trade-off of overall performance and forgetting on previous tasks.
The extensive experiments on four WSI datasets of different subtype classification tasks demonstrate the superiority of ConSlide in the effectiveness and better trade-off of overall performance and forgetting on previous tasks.
%The extensive ablation studies on four public WSI datasets validate the effectiveness of each proposed component and also the superiority of ConSlide over other methods under a fair WSI-based continual learning setting 
%
Moreover, the ConSlide is capable of managing survival prediction tasks by incorporating a Cox regression module.
We will investigate this potential application in the future.

\section{Acknowledgments}
% The work described in this paper was partially supported by grants from the National Natural Science Foundation of China (No. 62201483) and the Research Grants Council of the Hong Kong Special Administrative Region, China (No. 27206123 and T45-401/22-N).
The work described in this paper was partially supported by grants from the National Natural Science Foundation of China (No. 62201483) and the Research Grants Council of the Hong Kong Special Administrative Region, China (T45-401/22-N).

\if 0
% 
% Further, we devise a new asynchronous updating mechanism to encourage the patch-level and region-level blocks to learn task-agnostic and task-specific knowledge, respectively, by introducing a nested Self-Supervised Learning module.
Further, we devise a new asynchronous updating mechanism to encourage the patch-level and region-level blocks to learn low-level and high-level task-specific knowledge, respectively, by introducing a nested Self-Supervised Learning module.
% 
In the end, we perform ablation studies on four WSI datasets of different tasks to evaluate our method and compare it with other methods under a fair WSI-based continual learning setting to demonstrate its superiority in the effectiveness and better trade-off of overall performance and forgetting on previous tasks.
%
Furthermore, the ConSlide framework is capable of not only managing tumor subtype classification tasks but also incorporating the Cox regression to handle survival prediction tasks. We will investigate this potential application in the future.

% Our ASY-HIT performs best over other state-of-the-art methods with fair WSI-based continual learning setting and can better trade-off the overall performance and forgetting on previous tasks.
% % 
% And the extensive ablation study experiments show the effectiveness of our proposed HIT backbone design, asynchronous updating mechanism, BuRo rehearsal strategy, and synchronous mechanism.
\fi 

\clearpage

{\small
\bibliographystyle{ieee_fullname}
\bibliography{egbib}
}

\end{document}